\pdfoutput=1

\documentclass[11pt]{article}
\usepackage[table]{xcolor}
\usepackage[]{EMNLP2022}

\usepackage{times}
\usepackage{latexsym}

\usepackage[T1]{fontenc}

\usepackage[utf8]{inputenc}

\usepackage{microtype}
\usepackage{times}
\usepackage{pifont}
\usepackage{url}
\usepackage{algorithm}
\usepackage{algorithmic}
\usepackage{latexsym}
\usepackage{microtype}
\usepackage{graphicx}
\usepackage{subfigure}
\usepackage{pgf}
\usepackage{collcell}
\usepackage{booktabs}
\usepackage{amsmath}
\usepackage{amsthm}
\usepackage{enumitem}
\usepackage{amsfonts}
\usepackage{color}
\usepackage{multirow}
\usepackage{tikz}
\usepackage{pgfplots}
\usepackage{cleveref}
\usepackage{xspace}
\newcommand{\xmark}{\textcolor{red}{\ding{55}}}%
\newcommand{\cmark}{\textcolor{green}{\ding{52}}}%
\newcommand\tenc{\textsc{T-Enc}}
\newcommand\denc{\textsc{D-Enc}}
\newcommand\tdec{\textsc{T-Dec}}
\newcommand\ddec{\textsc{D-Dec}}

\newcommand\vx{\pmb{x}} \newcommand\vy{\pmb{y}}

\newcommand\vpsi{\pmb{\psi}}
\newcommand\vtheta{\pmb{\theta}}

\def\ie{{\em i.e.,}\xspace}

\def\tabref#1{Table~\ref{#1}}

\def\Secref#1{Section~\ref{#1}}

\def\eqref#1{(\ref{#1})}

\usepackage{highlight}
\renewcommand{\opacity}{50}
\usepackage{inconsolata}

\title{Can Domains Be Transferred Across Languages in Multi-Domain Multilingual Neural Machine Translation?}

\author{Thuy-Trang Vu$^\diamondsuit$\Thanks{ Work done while doing internship at eBay Inc.} \and Shahram Khadivi$^\dagger$ \\ \textbf{Xuanli He}$^\diamondsuit$  \and \textbf{Dinh Phung}$^\diamondsuit$ \and \textbf{Gholamreza Haffari}$^\diamondsuit$ \\
     $^\diamondsuit$Department of Data Science and AI, Monash University, Australia \\
     $^\dagger$ eBay Inc. \\
     \texttt{\{trang.vu1,xuanli.he1,first.last\}@monash.edu}\\\texttt{skhadivi@ebay.com}
      }

\begin{document}
\maketitle
\begin{abstract}
Previous works mostly focus on either multilingual or multi-domain aspects of neural machine translation (NMT). This paper investigates whether the domain information can be transferred across languages on the composition of multi-domain and multilingual NMT, particularly for the incomplete data condition where in-domain bitext is missing for some language pairs. 
Our results in the curated
leave-one-domain-out experiments show that multi-domain multilingual (MDML) NMT can boost zero-shot translation performance up to +10 gains on BLEU, as well as aid the generalisation of multi-domain NMT to the missing domain. 
We also explore  strategies for effective integration of multilingual and multi-domain NMT, including language and domain tag combination and auxiliary task training. We find that learning domain-aware representations 
and adding target-language tags to the encoder 
leads to effective MDML-NMT.
\end{abstract}

\section{Introduction}

Multilingual NMT (MNMT), which enables a single model to support translation across multiple directions, has  attracted a lot of interest both in the research community and industry. The gap between MNMT and bilingual counterparts has been reduced significantly, and even for some settings, it has been shown to surpass bilingual NMT~\cite{Tran2021FacebookAIWMT21}.
%
%
MNMT enables knowledge sharing among languages, and reduces model training, deployment, and maintenance costs. 
%
On the other hand, multi-domain NMT aims to build robust NMT models, providing high-quality translation on diverse domains. 
While multilingual and multi-domain NMT are highly appealing in practice, they are often studied separately. 

{To accommodate the domain aspect, previous MNMT works focus on learning a domain-specific MNMT by finetuning a general NMT model on the domain of interest~\cite{Tran2021FacebookAIWMT21,Berard2020MultilingualNeuralMachine}. Recently, \citet{Stickland2021MultilingualDomainAdaptation} 
propose to unify multilingual and multi-domain NMT into a holistic system by stacking language-specific and domain-specific adapters with a two-phase training process. Thanks to the plug-and-play ability of adapters, their system can handle translation across multiple languages and support multiple domains. However, as each domain adapter is learned independently, their adapter-based model lacks the ability of effective knowledge sharing among domains.}

\begin{figure}[]
    \centerline{\includegraphics[scale=0.223]{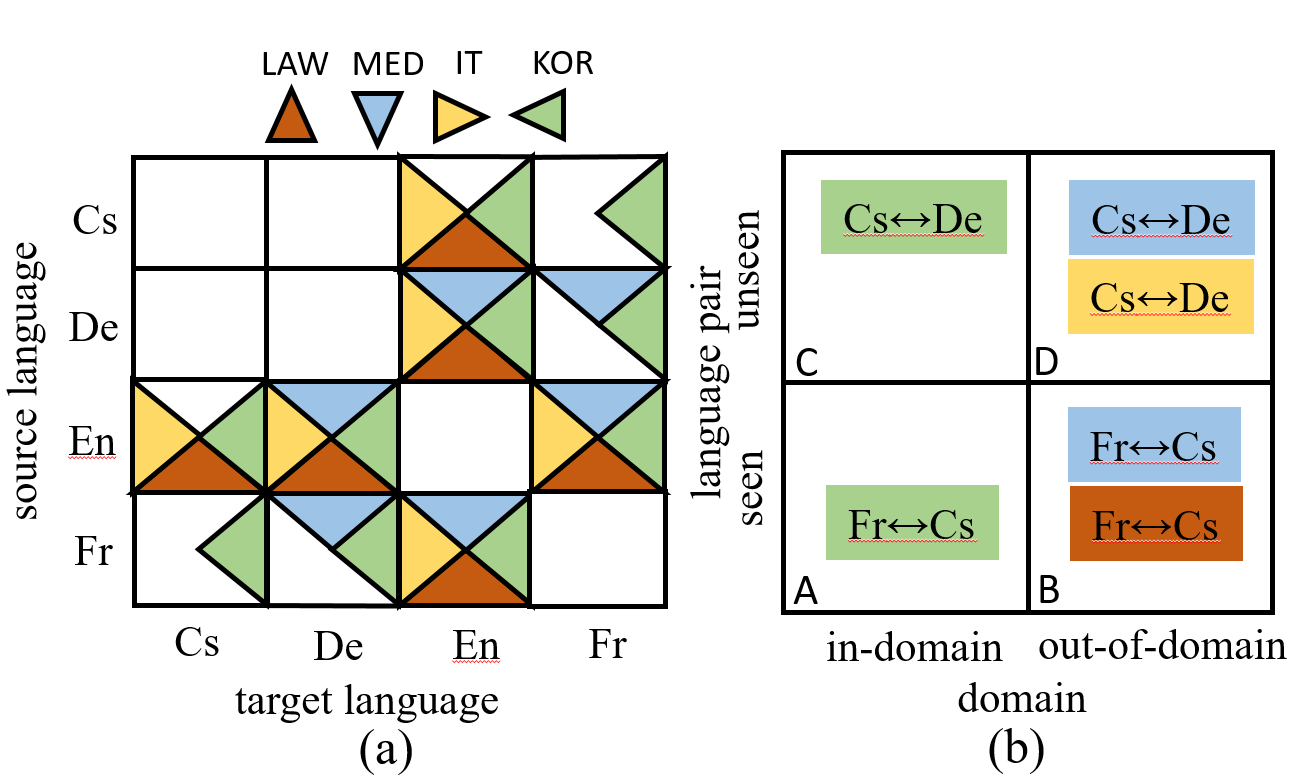}}
    \caption{{An example of the multi-domain multilingual incomplete data condition (best seen in colours). (a) The colour indicates  the availability of bitext in the corresponding domain for each language. (b) Domain and language-pair matrix for the data condition in (a).   
    }}
    \label{fig:data-condition}
\end{figure}

In this paper, we take a  step further toward unifying multilingual and multi-domain NMT into a single setting and model, \ie multi-domain multilingual NMT (MDML-NMT), {and enable effective knowledge sharing across both domains and languages.} Unlike the \emph{complete} data assumption in the multi-domain single language-pair setting where training data is available in all domains,
we assume the existence of bitext in all domains for only a subset of language-pairs, {as illustrated in~\Cref{fig:data-condition}(a).} In fact, it is highly improbable to obtain in-domain bitext for all domains and all language pairs in many real-life settings. 
{
Depending on the availability of parallel data, we categorise a translation task from a source  to a target language into four categories based on the following dimensions:
\begin{itemize}
    \item \emph{in-domain}/\emph{out-of-domain}, wrt to the domain of interest, and 
    \item \emph{seen}/\emph{unseen}, wrt to the translation direction during training.
\end{itemize}
Please note the domain and language-pair matrix in~\Cref{fig:data-condition}(b). In this figure,  parallel data available in the training set specifies the group A, the \emph{in-domain seen} tasks. Given this training dataset, most MNMT research focuses on cross-lingual transfer to \emph{in-domain unseen} translation tasks (A$\rightarrow$C), while the studies on multi-domain NMT and domain adaptation seek to generalise to \emph{out-of-domain seen} translation tasks (A$\rightarrow$B). Integrating domain and language aspects in the incomplete data condition gives rise to an interesting and more challenging setting that transfers to \emph{out-of-domain unseen} translation tasks (A$\rightarrow$D). We hypothesise that the out-of-domain ``seen and unseen'' translation tasks (A$\rightarrow$B+D) can benefit from the in-domain translation tasks if there exists the domain transfer across languages in MDML-NMT.
}

{Specifically, we ask the following research questions: (1) Do out-of-domain translation tasks benefit from the out-of-domain and in-domain bitext in other seen translation pairs? and (2) What is effective method to handle the composition of domains and languages? Furthermore, beyond the cross-lingual transfer (A$\rightarrow$C) and the out-of-domain generalisation (A$\rightarrow$B), we also consider the challenging setting where the translation direction of interest may not have any bitext in any domain, i.e. the zero-shot setting (A$\rightarrow$D).
}
{In general, we can vary the degree of domain transfer based on the number of domains in which parallel data for a translation task is available. Combining with the number of language pairs of interest, there are large numbers of incomplete data conditions, even for our toy examples in~\Cref{fig:data-condition}. In this study, we assume the highest degree of domain transfer and carefully design controlled experiments where one domain is left out for some language pairs (\Cref{tab:exp-setting}).
We then examine the potential of MDML-NMT on this incomplete data condition. We also explore training strategies for effective integration of multi-domain and multilingual NMT, mainly on (i) how to combine the language and domain tags, and (ii) using auxiliary task training to learn effective representations.} Our contributions are as follows:
\begin{itemize}
    \item 
    {We investigate effective  strategies to jointly learn multi-domain and multilingual NMT models under the incomplete data condition.}
    \item Our empirical results show that MDML-NMT model can improve translation quality in the zero-shot directions by mitigating the \textbf{off-target translation} issue that an MNMT model translates the input sentence to a wrong target language. 
     Additionally, MDML-NMT exhibits domain transfer ability by achieving up to +4 BLEU improvement over the multi-domain NMT
     on the translation direction where in-domain training data is absent. 
    Thanks to the effective cross-domain and cross-lingual knowledge sharing, MDML-NMT outperforms the adapter-based method \cite{Stickland2021MultilingualDomainAdaptation} by a large margin in the language-domain zero-shot setting.
    \item Our study sheds light on effective MDML-NMT training. 
    {Our experimental results reveal that: (i) for the domain, it is important to make the encoder domain-aware by either providing the domain tags or training with the auxiliary task;
    and (ii)  
    for the language, the best practice is to prepend the target language tag to the encoder.}
\end{itemize}

\begin{table}[]
\begin{center}
\scalebox{0.9}{
\begin{tabular}{l|ccccc}
\toprule
& \textsc{Law} & \textsc{IT} & \textsc{Koran} & \textsc{Med} & \textsc{Sub} \\
\midrule
\midrule
En-Fr & \cmark & \cmark & \cmark & \cmark & \cmark\\
En-De & \cmark & \cmark & \cmark & \cmark & \cmark\\
De-Fr & \cmark & \cmark & \cmark & \cmark & \cmark\\
En-Cs & \xmark & \cmark & \cmark & \cmark & \cmark\\
En-Pl & \xmark & \cmark & \cmark & \cmark & \cmark\\
\bottomrule
\end{tabular}
}
\caption{Illustration of leave-one-out \textsc{Law}  experiment setting. \xmark, \cmark describes whether there is bitext in the corresponding domain for the given language pairs.}
\label{tab:exp-setting}
\end{center}
\end{table}
\section{Multi-domain Multilingual NMT}
In this section, we first provide the necessary background on multilingual NMT (MNMT) and multi-domain NMT individually. We then describe effective modelling approaches for the integration of multi-domain and multilingual NMT (MDML-NMT).

\subsection{Multilingual NMT}
Given a set of languages $L$, the primary goal of MNMT is to learn a single NMT model that can handle all translation directions of interest in this set of languages~\citep{Dabre2020surveymultilingualneural}. 
According to the parameter sharing strategy, MNMT can be categorised into: 1) partial parameter sharing~\cite{Dong2015MultiTaskLearning,Firat2016MultiWayMultilingual,zhang2021share}, and 2) full parameter sharing~\cite{ha2016toward,Johnson2017GooglesMultilingualNeural}. The latter has been widely adopted because of its simplicity, lightweight, and its zero-shot capability. Thus, we adopt the full parameter sharing strategy in our work. 

In the fully parameter-shared MNMT, all parameters of encoders, decoders and attentions are shared across tasks. Special language tags are introduced to indicate the target languages. One can prepend the target language tags to either the source or target sentences. The model is then trained jointly to minimise the negative log-likelihood across all training instances: 
\begin{eqnarray}
\mathcal{L}_{\textrm{ML}}(\vtheta) :=- \sum_{(s,t)\in T}\sum_{(\vx,\vy) \in \mathcal{C}_{s,t}} \log P(\vy|\vx;\vtheta)
\label{eq:mnmt}
\end{eqnarray}
where $\vtheta$ is model parameters, $\mathcal{C}_{s,t}$ denotes a bilingual corpus for the source language $s$ and the target language $t$, $(\vx,\vy)$ is a pair of  parallel sentences in the source  and target language, and $T$ denotes the translation tasks for which we have bitext available. Among all possible language pairs $(s,t) \in L\times L$, we often only have access to bilingual data for a subset of them. We denote these pairs as \emph{seen} (observed) translation tasks,
and the rest as \emph{unseen} tasks corresponding to the zero-shot setting.

\subsection{Multi-domain NMT}
\label{sec:md-nmt}
Multi-domain NMT aims to handle translation tasks across multiple domains for a given language pair.  
Similar to MNMT, tagging the training corpus is the most popular approach, where a tag indicates the domain of a sentence pair. We also minimise the negative log-likelihood across all domains to train the model:
\begin{eqnarray}
\mathcal{L}_{\textrm{MD}}(\vtheta) :=-\sum_{d \in D}\sum_{(\vx,\vy) \in \mathcal{C}^{d}_{s,t}} \log P(\vy|\vx;\vtheta)
\label{eq:md-nmt}
\end{eqnarray}
where $D$ is the set of domains, and $\mathcal{C}^d_{s,t}$ denotes the parallel bitext in the source language $s$, target language $t$, and the domain $d$.

Apart from tagging, some auxiliary tasks have also been incorporated into the training process. 
A common practice is the use of domain discrimination, which aims to force the encoder to capture \textit{domain-aware} characteristics~\cite{Britz2017EffectiveDomainMixing}. For this purpose, a domain discriminator is added to the NMT model at training time. The input to the 
discriminator is the encoder output, and its output predicts the probability of the domain of the source sentence. 
The discriminator is jointly trained with the NMT model, and is discarded at inference time.

Let $\textbf{h} = \textrm{enc}(\vx)$ 
be the representation of sentence $\vx$ computed by the mean-pooling over the hidden states of the top layer of the encoder. The training objective for the domain-aware encoder is as follows:
\begin{eqnarray}
\mathcal{L}_{\textrm{disc}}(\vtheta, \vpsi) := - \displaystyle \sum_{d \in D}\sum_{(\vx,\vy) \in \mathcal{C}^{d}_{s,t}}\log \Pr(d|\textbf{h};\vpsi) \\
\mathcal{L}_{\textrm{MD-aware}}(\vtheta, \vpsi) :=  \mathcal{L}_{\textrm{MD}}(\vtheta) + \lambda \mathcal{L}_{\textrm{disc}}(\vtheta, \vpsi)
\label{eq:md-aware-nmt}
\end{eqnarray}
where $\vpsi$ is the parameter of the domain discriminator classifier, and $\lambda$ controls the contribution of the domain discriminator into the training objective of the multi-domain NMT model.


Alternatively, one can design an adversarial training objective in order to learn domain-agnostic representations by the encoder. This is achieved by inserting a gradient reversal layer~\cite{ganin2015unsupervised} between the encoder and the domain discriminator. The gradient reversal layer behaves as an identity layer in the forward pass but reverses the gradient sign during back-propagation. It has the opposite effect on the encoder, forcing it to learn domain-agnostic representations. This encourages the domain specific characteristic to be learned mainly by the decoder of the NMT model.

\subsection{Composition of Domains and Languages}
\label{sec:mdml}
In this paper, we explore strategies for composing multi-domain and multilingual NMT.
We consider the incomplete multi-domain multilingual data condition where in-domain data may be only available in a subset of language pairs. 
For example, Table \ref{tab:exp-setting} shows one of the data conditions explored in our experiments in \Secref{sec:experiments}. Given the five language pairs and five domains, we assume that the domain data in some language pairs are missing. Our goal is to investigate effective techniques to train a high-quality MDML-NMT model covering all combinations of domains and language pairs.  

\begin{table}[]
\begin{center}
\scalebox{0.73}{
\begin{tabular}{ccl}
\toprule
Trans. direction & Eval. domain & MDML task type \\
\midrule
En$\rightarrow$De & \textsc{Law} & seen in$\rightarrow$in \\
En$\rightarrow$Cs & \textsc{Law} & seen in$\rightarrow$out \\
Pl$\rightarrow$En & \textsc{Law} & seen out$\rightarrow$in \\
\midrule
De$\rightarrow$Cs & \textsc{Law} & unseen (zero-shot) in$\rightarrow$out \\
Cs$\rightarrow$De & \textsc{Law} & unseen (zero-shot) out$\rightarrow$in \\
Pl$\rightarrow$Cs & \textsc{Law} & unseen (zero-shot) out$\rightarrow$out \\
\bottomrule
\end{tabular}
}
\caption{Examples of MDML task types  in the leave-one-domain-out \textsc{Law} training scenario of  Table \ref{tab:exp-setting}. Please refer to~\tabref{tab:exp-setting} for the in/out and seen/unseen settings.}
\label{tab:task-eg}
\end{center}
\end{table}

\begin{figure*}[]
    \centerline{\includegraphics[scale=0.4]{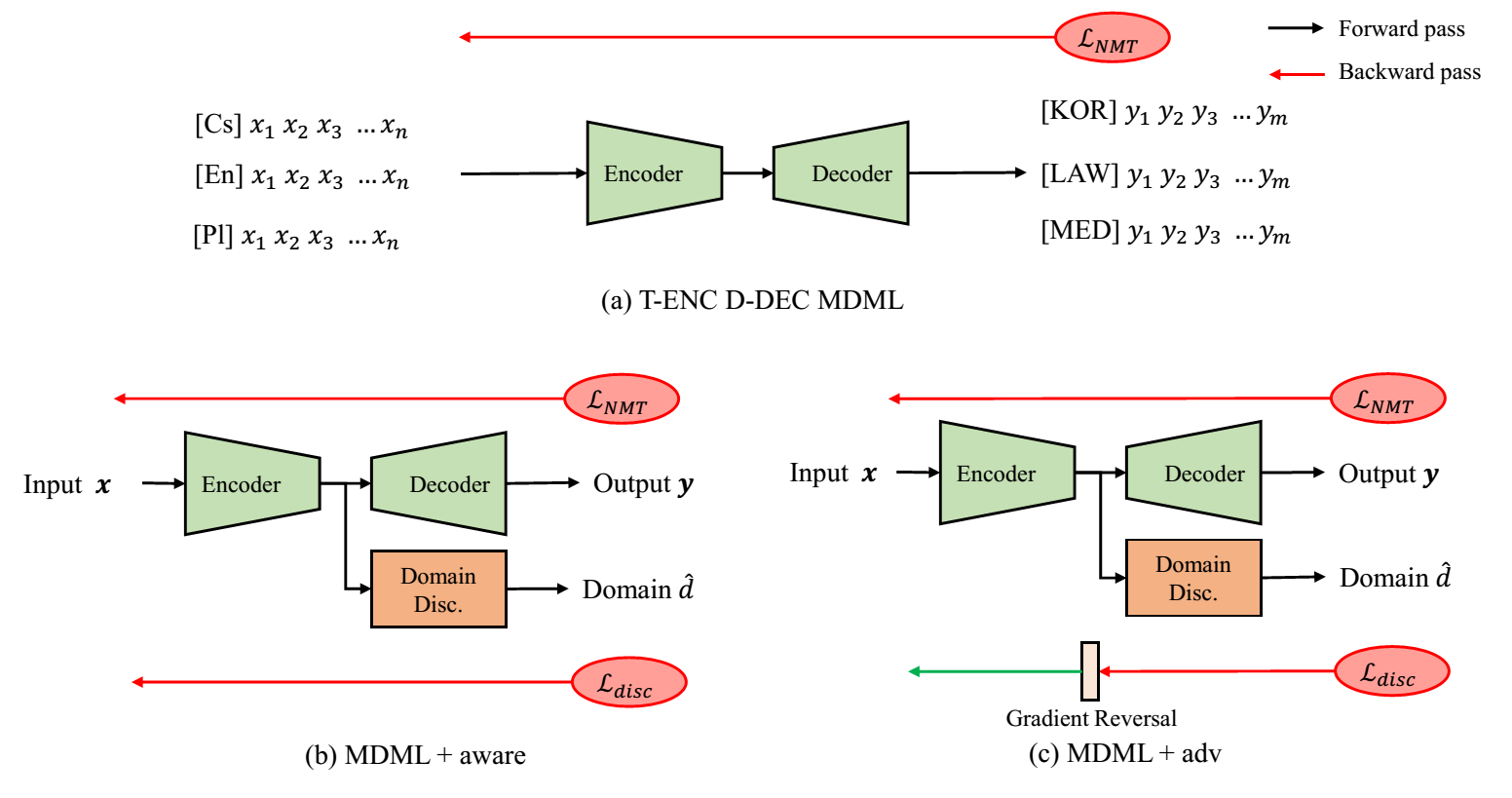}}
    \caption{{Illustration of domain and languages composition strategies: (a) prepending domain (D) and target language (T) tag to encoder (ENC) or decoder (DEC). This example shows a T-ENC D-DEC model where the target language tag and domain tag are added to encoder and decoder respectively; (b) combining the tagging method with the domain aware auxiliary task (MDML + aware) to learn domain-aware representation; and (c) combining the tagging method with the domain adversarial auxiliary task (MDML + adv) to learn domain-agnostic representation.}}
    \label{fig:mdml-model}
\end{figure*}

Given a specific domain, we define \textit{in-domain languages} as those having data available in the domain as part of some bilingual corpora; the rest of the languages are referred to as \textit{out-of-domain languages}. %
We consider all combinations of in-domain/out-of-domain source/target languages for both seen and unseen translation directions (see examples in Table \ref{tab:task-eg})  in  \Secref{sec:experiments}.


We investigate different combinations of the tagging strategy and auxiliary task training to effectively train MDML-NMT models, as shown in~\Cref{fig:mdml-model}.

\paragraph{Language and Domain Tags.}
We explore different ways of injecting the target language tags and domain tags into the translation process. Following the standard convention, we explore inserting the target language tag at the beginning of either the source sentence or the translation. Furthermore, the domain tag can also be added to either the source or the target side. 
\paragraph{Auxiliary Task Training.} 
We investigate the effect of encoder-based auxiliary tasks on MDML-NMT. As described in~\Secref{sec:md-nmt}, we consider  two types of auxiliary  objectives to train encoder which are domain-aware or domain-agnostic. The former aims to amplify the domain-related features, while the latter  focuses on the domain invariant representation in the encoder. 




\section{Experiments}
\label{sec:experiments}
In this section, we evaluate the MDML-NMT approaches and seek to answer the following research questions (\textbf{RQs}):
\begin{itemize}
    \item \textbf{RQ1}: 
    {\textit{Do out-of-domain translation tasks benefit from the out-of-domain and in-domain bitext in other translation pairs?}}
    
    {We explore the benefits of having a single MDML model trained on all available training data from multiple languages and domains over the multi-domain bilingual (MDBL) and the single domain multilingual (SDML) models learned on a subset of training data from a single language pair or domain. We carefully design controlled experiments to build incomplete data conditions and study the translation quality of the unified MDML-NMT model on both seen and unseen (zero-shot) translation directions. 
    We hypothesise that the translation involving the out-of-domain languages can be beneficial from the in-domain languages thanks to the knowledge sharing across domain and languages.}
    \item \textbf{RQ2}: \textit{What is effective method to handle composition of domains and languages?} 
    
    {We investigate strategies for effective integration of existing multi-domain and multilingual NMT methods, including the use of language and domain tags and auxiliary task training.}
\end{itemize}
\subsection{Setup}
We describe the experimental setup in this section, and then present our results. 
\paragraph{Dataset.}
We conduct experiments with translation  directions among five languages English (En), Czech (Cs), German (De), French (Fr) and Polish (Pl).
Following the recipe in \citet{Koehn2017SixChallengesNeural}, we create five domains: Law (\textsc{Law}) , IT (\textsc{IT}), Koran (\textsc{Kor}), Medical (\textsc{Med}), and Subtitles (\textsc{Sub}) from OPUS~\citep{tiedemann-2012-parallel}. 
These corpora are deduplicated and randomly selected, from each corpus 2K sentences extracted as the development and test sets in all possible translation pairs. The statistics of the training dataset are reported in
~\Cref{apdx:data}. 


\paragraph{Seen vs Unseen  Language Pairs.}
{We categorise the evaluated languages into two groups, high-resource languages including En, De, and Fr, for which bilingual data among these languages is easy to obtain.
We also consider low-resource languages, including Cs and Pl, for which only English-centric data is available, resulting in two language pairs. As a result, there are five \textit{seen} language pairs, consisting of ten seen translation directions.\footnote{This set consists of En-Fr, En-De, De-Fr, En-Cs, En-Pl.} There are also five \textit{unseen} language pairs, resulting in ten unseen translation directions; they are the ones for which we do not have any bitext in the dataset.\footnote{This set consists of De-Cs, De-Pl, Fr-Cs, Fr-Pl, Cs-Pl.}}

\paragraph{Leave-one-domain-out (LODO).} We {curate}
the incomplete MDML data condition by removing the data of one domain for the translations tasks involving low-resource languages. An example of the leave-one-domain-out data condition is shown in~\Cref{tab:exp-setting}.  
{In total, there are five LODO conditions, each of which corresponding to removing the bitext of one domain for both En-Cs and En-Pl (\ie our low-resource language pairs). For each of these LODO conditions, we have five seen language-pairs and  five unseen language-pairs, hence a total of 20 translation tasks in both directions.}

In the multi-domain NMT literature, this setting is {related} to domain generalisation which evaluates the NMT model on out-of-domain data in a zero-shot manner. By carefully removing only a specific domain, we would like to examine whether extra data (\ie the in-domain and out-of-domain data for high-resource languages, and out-of-domain data for low-resource languages) can boost the generalisation of MDML-NMT to the domain of interest.


\paragraph{Models.}
We use Transformer~\cite{Vaswani2017AttentionisAll} as the NMT model architecture and Fairseq implementation~\cite{Ott2019fairseqFastExtensible}. For all MDML-NMT models, we initialise them with mBART\_large~\cite{Liu2020MultilingualDenoisingPre}. We describe the model training details in~\Cref{apdx:training}.

As described in~\Cref{sec:mdml}, our approaches to MDML problem include combining language and domain tags, and adding domain auxiliary task to  the standard multilingual NMT objective. In the first approach, the target language tags can be inserted  to the source sentence (\textsc{T-Enc}) or the target sentence (\textsc{T-Dec}). The domain tags can also be handled in similar manners denoted as \textsc{D-Enc} and \textsc{D-Dec} respectively. On combining these tags, the language tag always appears first in the sentence. {In addition to the domain and language tag combination, we also explore whether learning domain-aware or domain-agnostic representation in the encoder with auxiliary task can aid MDML-NMT performance.~\Cref{fig:mdml-model} summarises the MDML-NMT approaches evaluated in this paper.}

{We also report the results of the adapter-based domain-specific MNMT, proposed by~\citet{Stickland2021MultilingualDomainAdaptation}. Language adapters~\cite{Bapna2019SimpleScalableAdaptation} are firstly injected to each layer of a pre-trained MNMT model and then trained while freezing the backbone. Then, domain adapters are stacked on top of the language adapters and trained without backpropagating to the MNMT backbone and the language adapters. Since we do not consider any additional parallel data apart from the multi-domain dataset, we train the MNMT backbone as well as the language and domain adapters using this multi-domain multilingual dataset (instead of Paracrawl) for fair comparison.}


\paragraph{Evaluation.} We report the detokenised BLEU scores calculated by SacreBLEU (Post, 2018)~\citep{Post2018CallClarityReporting} and the micro-average of BLEU score in a group as the measure of overall performance.\footnote{\url{nrefs:1|case:mixed|eff:no|tok:none|smooth:exp|version:2.0.0}}

\begin{table}[]
\begin{center}

\scalebox{1}{
\begin{tabular}{l|rr|rr}
\toprule
\multicolumn{1}{c}{}& \multicolumn{2}{c|}{\textsc{\denc}} & \multicolumn{2}{c}{\ddec} \\
\midrule
MDBL & \multicolumn{2}{c|}{12.43}  & \multicolumn{2}{c}{\underline{10.21}}\\
+adv & \multicolumn{2}{c|}{12.91} & \multicolumn{2}{c}{9.90}\\
+aware & \multicolumn{2}{c|}{\underline{13.13}} & \multicolumn{2}{c}{10.13}\\
\midrule 
\midrule
\multicolumn{1}{c}{}& \multicolumn{2}{c|}{\textsc{\denc}} & \multicolumn{2}{c}{\ddec} \\
\multicolumn{1}{c}{}& \multicolumn{1}{c}{\tenc} & \multicolumn{1}{c|}{\tdec} & \multicolumn{1}{c}{\tenc} & \multicolumn{1}{c}{\tdec} \\
\midrule
MDML & 14.48 & 13.21 & 14.11 & 8.16 \\
+adv & 14.91 & 14.30 & 14.72 & \underline{8.44} \\
+aware & \underline{15.00} & \underline{14.59} & \textbf{\underline{15.35}} & 7.99 \\
\bottomrule                         
\end{tabular}}
\caption{Average BLEU score of En$\rightarrow$Cs translation
across all leave-out domains for multi-domain multilingual (MDML) models and multi-domain bilingual (MDBL) models.
The best score on overall and within each tagging group are marked in \textbf{bold} and \underline{underline} respectively.
}
\label{tab:multidomain-sum}
\end{center}
\end{table}
\begin{table}[]
\begin{center}
\scalebox{1}{
\begin{tabular}{ll|rrr}
\toprule
 &  \multicolumn{1}{c}{} & \multicolumn{1}{c}{seen} & \multicolumn{1}{c}{unseen} & \multicolumn{1}{c}{unseen} \\
  & \multicolumn{1}{c}{} & \multicolumn{1}{c}{-both} & \multicolumn{1}{c}{-SDML} & \multicolumn{1}{c}{-both} \\
\midrule \midrule
\multirow{2}{*}{\tenc}& SDML & \textbf{41.40}  & 6.80 & 7.73  \\
& MDML & 37.25  & \textbf{21.72}  & \textbf{9.27}  \\
\midrule
\multirow{2}{*}{\tdec} & SDML & \textbf{41.03}  & 7.79 & 8.16 \\
& MDML & 35.44 & \textbf{21.43} & \textbf{14.73} \\
\bottomrule
\end{tabular}
}
\caption{Average BLEU scores of single-domain multilingual (SDML) and multi-domain multilingual (MDML) on the leave-out domains for three groups: (i) \emph{seen-both} - the three seen high-resource language pairs (En-De, En-Fr, De-Fr); (ii) \emph{unseen-SDML} - the two low-resource language pairs which are seen by MDML but unseen to SDML (En-Cs, En-Pl); and (iii) \emph{unseen-both} - the other five unseen language pairs.
}
\label{tab:sdml-sum}
\end{center}
\end{table}
\subsection{Results and Discussions}

\paragraph{Can multilinguality help the multi-domain learning? (MDBL vs. MDML)} {We first examine the potential of MDML over the counterpart multi-domain NMT model. \Cref{tab:multidomain-sum} shows the BLEU scores of MDBL and MDML for En$\rightarrow$Cs translation on various LODO settings. A breakdown of BLEU scores on leave-out domains is shown in~\Cref{tab:multidomain} in the Appendix~\ref{apx-addresults}. The
MDBL models are trained on all En$\rightarrow$Cs bilingual data except of the domain of interest.
Within the same tagging method, augmenting the NMT training with the domain auxiliary objectives (\ie domain-aware and domain-agnostic encoders) enhances the translation performance.}
The MDML models consistently surpass the corresponding MDBL settings, with an exceptional case, where both domain and language tags are applied to the decoder (\ie \tdec \ \ddec). This observation suggests there is knowledge sharing from in-domain languages to out-of-domain languages.

\begin{table}[]
\begin{center}
\scalebox{1}{
\begin{tabular}{l|rrr}
\toprule
\multicolumn{1}{c}{} & \multicolumn{1}{c}{MDML} & \multicolumn{1}{c}{+adv} & \multicolumn{1}{c}{+aware} \\
\midrule
\midrule
\tenc & 25.17 & 28.91 & \textbf{30.14} \\
\tenc\ \denc & 24.36  & 28.90 & \underline{29.23} \\
\tenc\ \ddec & 22.10 & 29.43 & \underline{29.94} \\
\tdec & 24.82 & 29.14 & \underline{29.52} \\
\tdec\ \denc & 24.95 & 28.56  & \underline{29.01}  \\
\tdec\ \ddec & \underline{19.19} & 17.68 & 14.37 \\
\midrule
Adapter-based & \multicolumn{3}{c}{23.26} \\
\bottomrule
\end{tabular}}
\caption{{Average BLEU score of MDML-NMT models across all five leave-one-out scenarios.}
The best score overall and within each tagging group are marked in \textbf{bold} and \underline{underline} respectively.
}
\label{tab:multilingual-sum}
\end{center}
\end{table}
\begin{table*}[]
\begin{center}
\scalebox{0.82}{   
\begin{tabular}{llrrr|rrr|r}
\toprule
                                   &                           & \multicolumn{3}{c}{seen (10)}                                                                               & \multicolumn{3}{|c}{unseen (zero-shot) (10)}    &\multicolumn{1}{|c}{ \multirow{2}{*}{AVG}}                                                                 \\
                        
\multicolumn{1}{c}{} & \multicolumn{1}{c}{ } & \multicolumn{1}{c}{in$\rightarrow$in (6)} & \multicolumn{1}{c}{in$\rightarrow$out (2)} & \multicolumn{1}{c}{out$\rightarrow$in (2)} & \multicolumn{1}{|c}{in$\rightarrow$out (4)} & \multicolumn{1}{c}{out$\rightarrow$in (4)} & \multicolumn{1}{c}{out$\rightarrow$out (2)}&\multicolumn{1}{|c}{} \\

\midrule
\midrule
\multicolumn{2}{c}{Adapter-based} & 34.32 & 11.76 & \textbf{33.34} & 7.38 & 6.86 & 6.84 & 16.75 \\
\midrule
\multirow{3}{*}{\tenc}             & MDML                      & 37.25                          & \textbf{14.63}                             & 29.05                             & 7.93                                & 10.35                               & 9.79 & 18.17 \\
                                   & +adv                  & 36.81                          & 13.88                             & 28.33                             & \textbf{10.91}                               & 22.05                               & 11.38 & 20.56 \\
                                   & +aware                & \underline{37.50}                          & 14.31                             & \underline{29.09}                             & 10.61                               & \underline{24.50}                               & \underline{11.94} & \underline{21.33}  \\
\midrule
\multirow{3}{*}{\shortstack[l]{\tenc\\ \denc}}       & MDML                      & 32.32                          & 11.52                             & 24.23                             & 7.22                                & 17.25                               & 7.85   & 16.73 \\
                                   & +adv                  & 37.24                          & \underline{13.66}                             & \underline{31.17}                             & \underline{10.20}                               & 24.21                               & \underline{11.67} & \textbf{21.36} \\
                                   & +aware                & \textbf{37.57}                          & 13.15                             & 31.15                             & 8.65                                & \underline{25.20}                               & 11.24 & 21.16 \\
\midrule
\multirow{3}{*}{\shortstack[l]{\tenc\\ \ddec}}       & MDML                      & 31.94                          & 10.55                             & 22.14                             & 5.88                                & 8.63                                & 5.83 & 14.16 \\
                                   & +adv                  & 36.70                          & \underline{12.85}                             & 25.38                             & \underline{10.61}                               & \underline{22.57}                               & \underline{9.52}  & \underline{19.61} \\
                                   & +aware                & \underline{37.47}                          & 12.08                             & \underline{25.59}                             & 10.08                               & 22.41                               & 9.01  & 19.44 \\
\midrule
\multirow{3}{*}{\tdec}             & MDML                      & 31.44                          & 11.25                             & 23.62                             & 7.63                                & 20.39                               & 8.59   & 17.15 \\
                                   & +adv                  & 36.92                          & 13.94                             & \underline{28.83}                             & 8.93                                & \underline{24.48}                               & 12.14  & 20.87 \\
                                   & +aware                & \underline{37.20}                          & \underline{14.00}                             & 28.62                             & \underline{10.30}                               & 23.95                               & \textbf{12.18} & \underline{21.04} \\
\midrule
\multirow{3}{*}{\shortstack[l]{\tdec\\ \denc}}       & MDML                      & 31.80                          & 10.40                             & 22.13                             & 5.97                                & 18.81                               & 7.47 & 16.10 \\
                                   & +adv                  & 36.35                          & 13.22                             & 27.96                             & 8.46                                & 24.35                               & 10.21  & 20.09  \\
                                   & +aware                & \underline{37.00}                          & \underline{13.32}                             & \underline{29.34}                             & \underline{9.57}                                & \textbf{25.89}                               & \underline{11.43}  & \underline{21.09} \\
\midrule
\multirow{3}{*}{\shortstack[l]{\tdec\\ \ddec}}       & MDML                      & \underline{30.17}                          & 4.77                              & 24.72                             & 3.65                                & 14.08                               & 4.43 & 13.64 \\
                                   & +adv                  & 25.18                          & \underline{6.04}                              & \underline{25.94}                             & \underline{5.88}                                & \underline{14.72}  & 6.37 & \underline{14.02} \\
                                   & +aware                & 20.61                          & 5.50                              & 23.27                             & 5.72                                & 7.64                                & \underline{6.40} & 11.52 \\
\bottomrule                                   
\end{tabular}
}
\caption{Average BLEU score on leave-out domain for different translation tasks. {We categorise 20 translation direction into \emph{seen} where the training data for the translation direction is available, otherwise \emph{unseen}. \emph{in} and \emph{out} show whether   the corresponding domain is observed during training or not (see \Cref{tab:task-eg} for a concrete example). The number in parentheses shows how many translation directions are in the corresponding category.} The best score of each column overall and within each tagging group are marked in \textbf{bold} and \underline{underline} respectively.
}
\label{tab:leave-out-domain}
\end{center}
\end{table*}
\paragraph{Can multi-domain data help multilingual NMT? (SDML vs. MDML)} {SDML models are domain-specific multilingual NMT models trained on the multilingual dataset in a given domain. As in-domain parallel data is absent for several language pairs, the MDML models are exposed to more seen translation tasks than SDML models thanks to the availability of out-of-domain data. Hence, for a given domain, we divide the evaluation  translation tasks into three groups: seen-both, unseen-SDML and unseen-both. The seen-both and unseen-both groups consist of translation directions which are observed and unobserved respectively by both models in training. The unseen-SDML group corresponds to those unseen by SDML, but seen by MDML models. We report the average performance of the MDML and SDML model on the leave-out domains in~\Cref{tab:sdml-sum}. The detailed results on each leave-out domain can be found in~\Cref{tab:sdml} in the Appendix~\ref{apx-addresults}. As expected, SDML  works well on the seen directions (seen-both) but behaves badly on the zero-shot settings (unseen-SDML and unseen-both). We speculate it is due to  the negative inference among domains. On the other hand, MDML outperforms SDML in unseen-SDML by a large margin thanks to the out-of-domain parallel data. Additionally, leveraging multi-domain data also helps to improve multilingual NMT on unseen-both tasks up to +6 BLEU score on average.}

\paragraph{What is an effective method to MDML?} 
{We have previously shown the benefits of MDML over multi-domain and multilingual NMT models. The remaining question is how to integrate the multi-domain and multilingual approaches effectively. We report the average BLEU scores of different MDML methods across all five LODO scenarios and 20 translation tasks in ~\Cref{tab:multilingual-sum}. Similar to the previous observation on En$\rightarrow$Cs translation, models with domain discriminator outperform the vanilla MNMT model in all tagging methods. More specifically, the domain-aware MNMT models (+aware) are the winning method in most scenarios. These results emphasise the importance of having domain-aware representation in the encoder. Furthermore, it shows MDML is more effective than the adapter-based approach.}
 

\begin{table*}[t]
\begin{center}

\scalebox{0.65}{
\begin{tabular}{llrrrrr|rrrr}
\toprule
                             &            & \multicolumn{5}{c}{seen}                                                                                                   & \multicolumn{4}{|c}{unseen (zeroshot)}                                                                        \\
                             &            & \multicolumn{1}{c}{En} & \multicolumn{1}{c}{De} & \multicolumn{1}{c}{Fr} & \multicolumn{1}{c}{Cs} & \multicolumn{1}{c}{Pl} & \multicolumn{1}{|c}{De} & \multicolumn{1}{c}{Fr} & \multicolumn{1}{c}{Cs} & \multicolumn{1}{c}{Pl} \\
\midrule
\multirow{3}{*}{\tenc} & MDML & \cca{ 94.72} & \cca{ 95.99} & \cca{ 95.54} & \cca{ 92.10} & \cca{ 94.50} & \cca{ 48.66} & \cca{ 49.38} & \cca{ 32.73} & \cca{ 40.78} \\
 & +adv & \cca{ 94.81} & \cca{ 96.01} & \cca{ 95.33} & \cca{ 91.62} & \cca{ 95.06} & \cca{ 75.56} & \cca{ 85.93} & \cca{ 59.18} & \cca{ 66.57} \\
 & +aware & \cca{ 94.85} & \cca{ 96.09} & \cca{ 95.56} & \cca{ 91.60} & \cca{ 94.69} & \cca{ 80.92} & \cca{ 90.86} & \cca{ 64.77} & \cca{ 74.67} \\
\midrule
\multirow{3}{*}{\shortstack[l]{\tenc\\ \denc}} & MDML & \cca{ 92.55} & \cca{ 95.54} & \cca{ 95.06} & \cca{ 91.21} & \cca{ 94.12} & \cca{ 73.99} & \cca{ 72.85} & \cca{ 44.69} & \cca{ 58.83} \\
 & +adv & \cca{ 94.65} & \cca{ 96.11} & \cca{ 95.42} & \cca{ 90.51} & \cca{ 93.60} & \cca{ 80.30} & \cca{ 81.16} & \cca{ 59.13} & \cca{ 67.17} \\
 & +aware & \cca{ 94.67} & \cca{ 96.18} & \cca{ 95.44} & \cca{ 90.35} & \cca{ 92.69} & \cca{ 81.21} & \cca{ 84.05} & \cca{ 61.10} & \cca{ 66.92} \\
\midrule
\multirow{3}{*}{\shortstack[l]{\tenc\\ \ddec}} & MDML & \cca{ 94.33} & \cca{ 95.22} & \cca{ 95.10} & \cca{ 91.26} & \cca{ 94.98} & \cca{ 43.53} & \cca{ 46.94} & \cca{ 36.10} & \cca{ 44.04} \\
 & +adv & \cca{ 94.86} & \cca{ 95.81} & \cca{ 95.44} & \cca{ 91.55} & \cca{ 94.64} & \cca{ 87.23} & \cca{ 90.67} & \cca{ 69.01} & \cca{ 74.42} \\
 & +aware & \cca{ 95.01} & \cca{ 96.01} & \cca{ 95.49} & \cca{ 91.34} & \cca{ 94.46} & \cca{ 82.00} & \cca{ 91.38} & \cca{ 68.75} & \cca{ 75.69} \\
\midrule
\multirow{3}{*}{\tdec} & MDML & \cca{ 94.03} & \cca{ 95.32} & \cca{ 95.04} & \cca{ 90.70} & \cca{ 93.83} & \cca{ 90.44} & \cca{ 92.74} & \cca{ 60.44} & \cca{ 70.93} \\
 & +adv & \cca{ 94.68} & \cca{ 96.05} & \cca{ 95.44} & \cca{ 91.72} & \cca{ 94.84} & \cca{ 86.03} & \cca{ 88.64} & \cca{ 52.45} & \cca{ 64.01} \\
 & +aware & \cca{ 94.72} & \cca{ 96.21} & \cca{ 95.50} & \cca{ 92.22} & \cca{ 95.13} & \cca{ 77.20} & \cca{ 87.61} & \cca{ 58.17} & \cca{ 70.86} \\
\midrule
\multirow{3}{*}{\shortstack[l]{\tdec\\ \denc}} & MDML & \cca{ 92.72} & \cca{ 95.51} & \cca{ 95.06} & \cca{ 89.72} & \cca{ 92.23} & \cca{ 85.75} & \cca{ 89.82} & \cca{ 56.74} & \cca{ 68.56} \\
 & +adv & \cca{ 93.82} & \cca{ 96.14} & \cca{ 95.53} & \cca{ 91.41} & \cca{ 94.27} & \cca{ 84.70} & \cca{ 87.99} & \cca{ 51.59} & \cca{ 63.66} \\
 & +aware & \cca{ 94.19} & \cca{ 96.12} & \cca{ 95.54} & \cca{ 91.54} & \cca{ 93.80} & \cca{ 79.93} & \cca{ 87.00} & \cca{ 60.22} & \cca{ 72.77} \\
\midrule
\multirow{3}{*}{\shortstack[l]{\tdec\\ \ddec}} & MDML & \cca{ 93.44} & \cca{ 90.30} & \cca{ 93.44} & \cca{ 74.49} & \cca{ 83.06} & \cca{ 64.33} & \cca{ 58.96} & \cca{ 21.42} & \cca{ 25.74} \\
 & +adv & \cca{ 80.29} & \cca{ 17.71} & \cca{ 94.36} & \cca{ 49.29} & \cca{ 16.07} & \cca{ 3.37 } & \cca{ 47.72} & \cca{ 1.04 } & \cca{ 0.62 } \\
 & +aware & \cca{ 69.89} & \cca{ 14.25} & \cca{ 85.62} & \cca{ 52.34} & \cca{ 10.10} & \cca{ 2.89 } & \cca{ 9.52 } & \cca{ 2.07 } & \cca{ 0.28} \\
\bottomrule
\\
&&&\multicolumn{1}{c}{\cca{0}} & \multicolumn{1}{c}{\cca{25}}&\multicolumn{1}{c}{\cca{50}}&\multicolumn{1}{c}{\cca{75}}&\multicolumn{1}{c}{\cca{100}}&&&
\end{tabular}}
\caption{On-target translation ratio of MDML-NMT models on the seen and unseen translation tasks.}
\label{tab:on-target}
\end{center}
\end{table*}

As illustrated in~\Cref{tab:task-eg}, translation tasks in MDML setting can be categorised into seen and unseen (zero-shot) tasks involving the in-domain or out-of-domain languages.~\Cref{tab:leave-out-domain} reports the performance of MDML-NMT models in the leave-out domains on different task categories, e.g. \textsc{Law} in the example in~\Cref{tab:exp-setting}. The results for other domains, i.e. excluding the leave-out domains, can be found in~\Cref{apx-addresults}. Consistent with previous findings, the domain discriminative mixing methods outperform the other models. While the best multilingual NMT model (MDML \textsc{T-Enc}) performs comparably with other MDML-NMT models on seen translation tasks, the main benefit of MDML-NMT models comes from unseen translation tasks. 
As expected, for both seen and unseen tasks, the quality of translation when translating into in-domain languages is consistently higher than  
into out-of-domain languages.
{Stacking the language and domain adapters works particularly well in seen translation direction to in-domain target languages. Aligned with previous findings, the adapter-based method struggles to translate to out-domain target languages due to the unobserved combination of language and domain adapters during training~\citep{Stickland2021MultilingualDomainAdaptation}.} 

\section{Analysis}
\subsection{Domain-specific token generation}

In this section, we will look at how well MDML models are in generating domain-specific tokens. We concatenate all training data in a given domain in each language, remove stopwords, and extract the top 1000 domain-specific tokens with TF-IDF. The stopwords for each language are obtained from stopwords-iso\footnote{\url{https://github.com/stopwords-iso/stopwords-iso}}. 
\Cref{tab:domain-token} reports the F1 score of MDML models in generating leave-out domain-specific tokens. As expected, translation to in-domain languages (in$\rightarrow$in, out$\rightarrow$in) has a higher F1 score than translation to out-of-domain languages (in$\rightarrow$out, out$\rightarrow$out). Compared to MDML, both MDML-aware and MDML-adv models are able to generate more domain-specific tokens.

\begin{table}[t]
\begin{center}
\scalebox{0.7}{
\begin{tabular}{llllll}
\toprule
& & \multicolumn{1}{c}{in$\rightarrow$in}  &  \multicolumn{1}{c}{in$\rightarrow$out}  &  \multicolumn{1}{c}{out$\rightarrow$in}  &  \multicolumn{1}{c}{out$\rightarrow$out} \\
\midrule
\multirow{3}{*}{\tenc}       &  MDML                  & \cca{ 63.22 } & \cca{ 21.45   } & \cca{ 35.58   } & \cca{ 16.42    }\\
                              &  +adv    & \cca{ 62.71 } & \cca{ 26.73   } & \cca{ 44.48   } & \cca{ 22.06    }\\
                              &  +aware  & \cca{ 63.45 } & \cca{ 25.85   } & \cca{ 47.53   } & \cca{ 23.96    }\\
\midrule
\multirow{3}{*}{\shortstack[l]{\tenc\\ \denc}}  &  MDML                  & \cca{ 58.93 } & \cca{ 20.58   } & \cca{ 35.17   } & \cca{ 16.10    }\\
                              &  +adv    & \cca{ 63.14 } & \cca{ 24.24   } & \cca{ 46.75   } & \cca{ 23.55    }\\
                              &  +aware  & \cca{ 63.48 } & \cca{ 20.80   } & \cca{ 47.82   } & \cca{ 23.17    }\\
\midrule
\multirow{3}{*}{\shortstack[l]{\tenc\\ \ddec}}  &  MDML                  & \cca{ 58.82 } & \cca{ 20.32   } & \cca{ 30.34   } & \cca{ 13.37    }\\
                              &  +adv    & \cca{ 62.83 } & \cca{ 27.68   } & \cca{ 47.02   } & \cca{ 26.21    }\\
                              &  +aware  & \cca{ 63.59 } & \cca{ 27.64   } & \cca{ 47.21   } & \cca{ 25.73    }\\
\midrule
\multirow{3}{*}{\tdec}       &  MDML                  & \cca{ 58.35 } & \cca{ 21.70   } & \cca{ 43.69   } & \cca{ 19.98    }\\
                              &  +adv    & \cca{ 62.83 } & \cca{ 23.72   } & \cca{ 47.55   } & \cca{ 25.56    }\\
                              &  +aware  & \cca{ 63.08 } & \cca{ 25.90   } & \cca{ 47.31   } & \cca{ 25.49    }\\
\midrule
\multirow{3}{*}{\shortstack[l]{\tdec\\ \denc}}  &  MDML                  & \cca{ 58.94 } & \cca{ 18.34   } & \cca{ 38.85   } & \cca{ 17.56    }\\
                              &  +adv    & \cca{ 62.37 } & \cca{ 21.86   } & \cca{ 45.67   } & \cca{ 20.31    }\\
                              &  +aware  & \cca{ 62.98 } & \cca{ 24.23   } & \cca{ 47.87   } & \cca{ 24.17    }\\
\midrule
\multirow{3}{*}{\shortstack[l]{\tdec\\ \ddec}}  &  MDML                  & \cca{ 56.74 } & \cca{ 12.52   } & \cca{ 40.01   } & \cca{ 10.98    }\\
                              &  +adv    & \cca{ 46.71 } & \cca{ 9.18    } & \cca{ 34.73   } & \cca{ 8.34     }\\
                              &  +aware  & \cca{ 39.20 } & \cca{ 8.93    } & \cca{ 26.32   } & \cca{ 8.06 }\\   
\bottomrule
\\
&\multicolumn{1}{c}{\cca{0}} & \multicolumn{1}{c}{\cca{25}}&\multicolumn{1}{c}{\cca{50}}&\multicolumn{1}{c}{\cca{75}}&\multicolumn{1}{c}{\cca{100}}
\end{tabular}
}
\caption{In-domain token generation F1 score.}
\label{tab:domain-token}
\end{center}
\end{table}

\begin{table}[]
\begin{center}
\scalebox{0.65}{
\begin{tabular}{lcrrrrr}
\toprule
& \multicolumn{1}{c}{} & \multicolumn{1}{c}{En} & \multicolumn{1}{c}{De} & \multicolumn{1}{c}{Fr} & \multicolumn{1}{c}{Cs} & \multicolumn{1}{c}{Pl} \\
\midrule
\multirow{5}{*}{\shortstack[l]{LO}}  & En & & \cca{ 91.60} & \cca{ 93.27} & \cca{ 33.30} & \cca{ 43.99} \\
& De & \cca{ 93.26} &  & \cca{ 90.79} & \cca{ 9.88 } & \cca{ 10.82} \\
& Fr & \cca{ 90.59} & \cca{ 83.86} &  & \cca{ 2.53 } & \cca{ 4.90 } \\
& Cs & \cca{ 91.51} & \cca{ 68.39} & \cca{ 64.78} &  & \cca{ 7.87 } \\
& Pl & \cca{ 92.18} & \cca{ 66.58} & \cca{ 64.96} & \cca{ 18.83} &  \\
\midrule
\multirow{5}{*}{others} & En &  & \cca{ 95.38} & \cca{ 94.79} & \cca{ 84.79} & \cca{ 92.83} \\
& De & \cca{ 94.58} &  & \cca{ 92.78} & \cca{ 37.27} & \cca{ 46.95} \\
& Fr & \cca{ 91.70} & \cca{ 86.50} &  & \cca{ 22.04} & \cca{ 28.20} \\
& Cs & \cca{ 94.37} & \cca{ 63.79} & \cca{ 58.49} & & \cca{ 15.48} \\
& Pl & \cca{ 94.66} & \cca{ 63.29} & \cca{ 56.46} & \cca{ 13.21} & \\   
\bottomrule
\\
&&\multicolumn{1}{c}{\cca{0}} & \multicolumn{1}{c}{\cca{25}}&\multicolumn{1}{c}{\cca{50}}&\multicolumn{1}{c}{\cca{75}}&\multicolumn{1}{c}{\cca{100}}
\end{tabular}
}
\caption{On target ratio of \tdec \ \ddec \ MDML on the leave-out (LO) and other domains. Rows and columns correspond to the source and target languages.}
\label{tab:leave-out-on-tgt-ratio}
\end{center}
\end{table}
\subsection{On-target translation ratio} 

  \begin{figure*}[]
    \centerline{\includegraphics[scale=0.33]{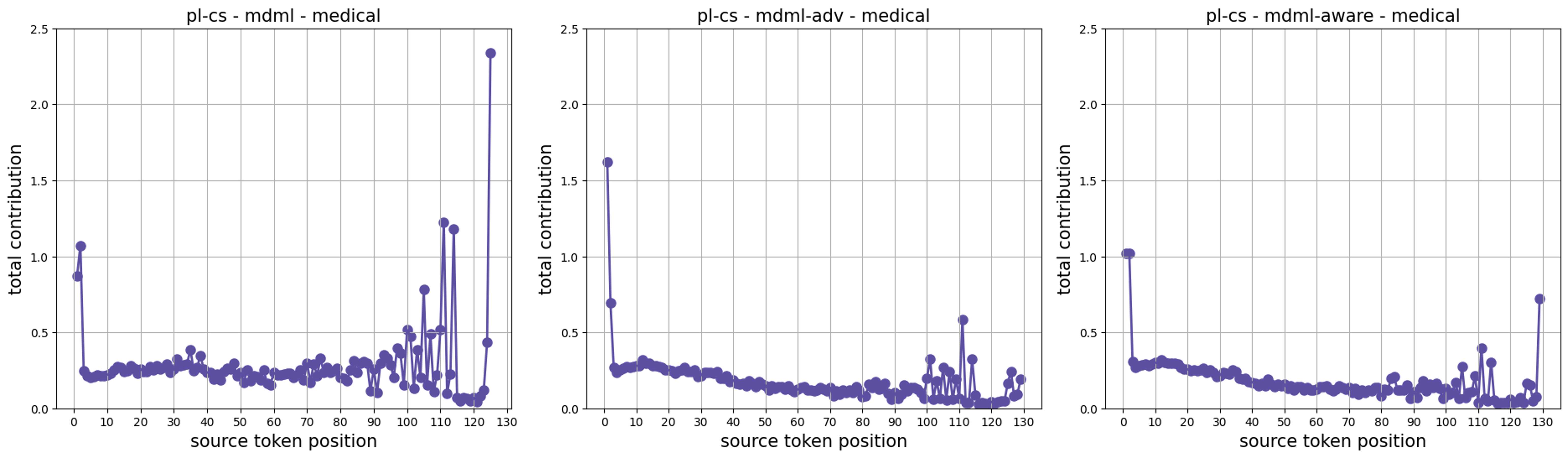}}
    \caption{Source token contribution on Pl$\rightarrow$Cs MDML with \tenc \ \denc. The target language and domain tag are the first two tokens.}
    \label{fig:lrp-language-tag-en-cs}
\end{figure*}

One challenge of multilingual NMT (MNMT) is the off-target translation in zero-shot direction. Off-target translation is an issue that the MNMT model translates the input sentence to the wrong language, causing low BLEU scores. In this section, we assess the ability to alleviate the off-target issue in MDML models. ~\Cref{tab:on-target} reports the on-target translation ratio of MDML models on seen and unseen translation for different target languages. We detect the language of translated targets using langdetect\footnote{\url{https://github.com/Mimino666/langdetect}} tool and calculate the on-target translation ratio as the percentage of translated sentences having the target language detected correctly. As expected, the seen translation tasks have more than 90\% sentences in the correct target language, except \tdec \ \ddec~models. On the other hand, the unseen tasks suffer from a low ratio, especially for Cs and Pl. We also observe significant improvement from MDML-aware and MDML-adv over the MDML models on unseen translation tasks to Cs and Pl.

Generally, \tdec \ \ddec~model always underperforms other models and have a much lower on-target ratio on unseen tasks. ~\Cref{tab:leave-out-on-tgt-ratio} further confirms this phenomenon on the leave-out domains.  While heavily suffering from the off-target issue in the leave-out domains, it has comparable ratios to other methods in other domains on seen tasks En-Pl and En-Cs. One possible explanation is that the combination of the target language and domain tags has never been observed during training for the unseen tasks with out-of-domain languages. 


\subsection{Language and domain tag contribution}
 To understand the role of the target and language tags to the generated prediction, we estimate the total contribution of source tokens at each position to the whole target sentence using Layerwise Relevance Propagation \citep{Voita2021AnalyzingSourceTarget}. We filter out the pairs having too short or too long target sentences and compute the contribution to target sentences of length between 10 and 100.  
 
 Results of \tenc \ \denc \ MDML models on Pl$\rightarrow$Cs translation in the leave-out medical domain are shown in ~\Cref{fig:lrp-language-tag-en-cs}. The language and domain tag are the first two source tokens in respective order. It can be seen that all models have a similar trend in which the contribution of source tokens decreases toward later positions and suddenly increases at a few last positions. Additionally, the target language tags play an important role in the final prediction of all MDML models. Interestingly, while still having a fairly high contribution compared to other tokens, the domain tag seems less important for the domain adversarial models. It can be explained that the encoder learns to produce domain agnostic representation; hence less depends on the domain tags.

\section{Related works}
\paragraph{Multilingual NMT.}
As a remarkable branch of NMT, multilingual NMT (MNMT) has been appealing for its capability of supporting translations among different language pairs. 
\citet{Dong2015MultiTaskLearning} opened the door to the MNMT by conducting a one-to-many translation. \citet{Firat2016MultiWayMultilingual} effectively extend this approach to a many-to-many setting. Since these approaches consider each translation as an independent system, they suffer from two major drawbacks. First, as the parameter size is proportional to the language size, it is not parameter-efficient when scaling to tens or hundreds of languages. In addition, the separate architectures cannot fully benefit from cross-lingual knowledge transfer.
\citet{Johnson2017GooglesMultilingualNeural,ha2016toward} devise a universal MNMT system to alleviate these issues by prepending a target language tag to the inputs and training a shared \textsc{Seq2Seq} model on the concatenation of all bitext. However, owing to the negative interference, high-resource languages suffer from translation inferiority, compared to the corresponding bilingual NMT models. 
As a remedy, ~\citet{zhang2021share, kudugunta2021beyond} leverage a mixture-of-experts design to separate language-specific features from the generic features by incorporating language-specific components into the universal MNMT model. Besides, \citet{Bapna2019SimpleScalableAdaptation,Zhu2021SerialParallel?Pluga} propose to fine-tune a lightweight adapter as a means of compensation for the quality loss caused by the adverse effect.

\paragraph{Multidomain NMT.} While both involving training on dataset coming from multiple domains, NMT domain adaption is different from multi-domain NMT. The former aims to transfer the knowledge of out-of-domain data into the in-domain data~\cite{luong-manning-2015-stanford,zoph2016transfer,freitag2016fast}, while the latter focuses on building a system, performing well on multiple domains~\citep{Pham2021RevisitingMultiDomain}.
Since lexical and topic variations have been observed in different domains, it is challenging to handle the mixed-domain data with a generic NMT model~\cite{farajian-etal-2017-neural}. 
To operate translation in multiple domains, 
recent research focuses on exploiting domain-shared and domain-specific knowledge by introducing a domain tag to the source sentence~\cite{KOBUS2017DomainControlNeural}, using auxiliary objectives such as domain discrimination loss~\cite{Britz2017EffectiveDomainMixing, Gu2019ImprovingDomainAdaptation}, domain knowledge distillation~\citep{Currey2020DistillingMultipleDomains}, and modifying the architecture to capture this information explicitly~\citep{Zeng2018MultiDomainNeural}. Rather than using a heavy domain-specific encoder-decoder architecture, \citet{Wang2020Gogeneralparticular} introduce lightweight domain transformation layers between the shared encoder and decoder. 


\paragraph{Multilingual \& multi-domain NMT.} Previous works have mainly considered multilingual and multi-domain NMT models as two disjoint systems. Until recently, \citet{Stickland2021MultilingualDomainAdaptation} propose to unify these two settings into a holistic system, but focus more on the domain adaptation angle. They investigate the combination of language and domain adapters by superimposing domain adapters on language adapters. 
They noticed that domain adapters and back-translation could boost the translation quality on the out-of-domain languages. 
In contrast, our work creatively stitches multilingual and multi-domain NMT together and explores the capability of a cross-lingual domain transfer within a unified model without adaption.




\section{Conclusion}

We study the problem of MDML-NMT for which a single NMT can support multiple translation directions and domains. We investigate whether the tagging and auxiliary task learning method can be combined for MDML-NMT.
Our empirical results reveal a positive transfer from in-domain to out-of-domain languages, especially in the zero-shot scenario. 
This study provides insights into the synergy of the domain and language aspects of training an MDML-NMT model. The main findings include: (i) it is crucial to make the encoder domain-aware; and (ii) it is best to prepend the target language tag to the encoder in MDML. These findings lay the groundwork for future research in this direction.

\section*{Acknowledgments}
This research is supported by an eBay Research Award and the ARC Future Fellowship FT190100039. This work is partly sponsored by the Air Force Research Laboratory and DARPA under agreement number FA8750-19-2-0501. The U.S. Government is authorized to reproduce and distribute reprints for Governmental purposes notwithstanding any copyright notation thereon. The  authors  are  grateful  to Minghao Wu, Michelle Zhao and the  anonymous  reviewers for their helpful comments to improve the manuscript.
\bibliographystyle{acl_natbib}
\bibliography{mdml}

\clearpage
\appendix

\section{Data statistics}
\label{apdx:data}
\Cref{tab:dataset} shows the statistics of dataset used in the experiments.
\begin{table}[]
\begin{center}
\scalebox{0.9}{
\begin{tabular}{l||rrrrr}
\toprule
\textbf{Domain} & \multicolumn{1}{c}{\textbf{Cs-En}}  &\multicolumn{1}{c}{\textbf{De-En}} & \multicolumn{1}{c}{\textbf{Fr-En}}  &\multicolumn{1}{c}{\textbf{Pl-En}} & \multicolumn{1}{c}{\textbf{De-Fr}} \\
\midrule
\textsc{Law} & 1.3M  & 467K & 596K  & 1.0M & 1.3M \\
\textsc{IT}  & 73K  & 158K  & 230K  & 97K & 146K \\
\textsc{Med} & 686K & 705K   & 705K  & 666K & 707K \\
\textsc{Kor} & 117K & 17.8K & 28K  & 30K & 10K \\
\textsc{Sub} & 595K & 494K & 492K & 491K & 590K \\
\bottomrule                 
\end{tabular}
}
\caption{Number of training sentences in the evaluation datasets. Each dataset contains 2K dev and test sentences.}
\label{tab:dataset}
\end{center}
\vspace{-5mm}
\end{table}

\section{Training Details}
\label{apdx:training}
For all MDML-NMT models, we initialise them with mBART\_large~\cite{Liu2020MultilingualDenoisingPre} and train with mixed-precision training up to 200K update steps (around 13 epochs) using a batch size of 8192 tokens and early stopping on 8 V100 GPUs. The multi-domain NMT (MDBL) is trained in a similar manner, except with the total update steps of 60K which is equivalent to around 30 epochs. We apply Adam with an inverse square root schedule, a linear warmup of 5000 steps and a learning rate of 3e-5. We set dropout and label smoothing with a rate of 0.3 and 0.2. We use temperature-based sampling with $T=5$ to balance training size between domains and languages~\citep{Arivazhagan2019Massivelymultilingualneural}.

For the NMT model with auxiliary task, the domain discriminator is a 2-layer feed-forward network with hidden size of 1024. We set the mixing hyperparameters $\lambda$ in Equation~\ref{eq:md-aware-nmt} to 1, \ie the domain discriminative loss and NMT loss contributes equally to the training signal.

Followed~\citep{Stickland2021MultilingualDomainAdaptation}, we use adapter bottle-neck of 1024 for the adapter-based models. The monolingual language adapters are trained all together on the multi-domain dataset while the NMT backbone are frozen. In contrast, we train domain adapters separately for each domain and build homogeneous batches containing sentences from the same language direction and domain. We also apply domain-adapter dropout (DADrop) where the domain adapters are skipped 20\% of time.

\section{Additional Results}
\label{apx-addresults}
\paragraph{MDBL vs. MDML.} \Cref{tab:multidomain} shows the BLEU scores of different models for En$\rightarrow$Cs translation on various LODO settings. Each domain column reports the results corresponding to the LODO setting in which the bitext of that domain is removed.

\paragraph{SDML vs. MDML.} We report the performance of the MDML and SDML model on each leave-out domains in~\Cref{tab:sdml}.

\paragraph{MDML Result.} The average BLEU scores on each domain across all five LODO scenarios and 20 translation tasks are reported in~\Cref{tab:multilingual}. \Cref{tab:other-domain} reports the performance of MDML-NMT models on other domains (excluding the leave-out domains) on different task categories.
\begin{table*}[]
\begin{center}

\scalebox{1}{
\begin{tabular}{ll|rrrrr|r}
\toprule
& & \multicolumn{1}{c}{\textsc{Law}} & \multicolumn{1}{c}{\textsc{IT}} & \multicolumn{1}{c}{\textsc{Kor}} & \multicolumn{1}{c}{\textsc{Med}} & \multicolumn{1}{c}{\textsc{Sub}} & \multicolumn{1}{c}{AVG}  \\
\midrule
\midrule
\multirow{3}{*}{\denc}      & MDBL         & 10.52 & 20.37 &  7.63 & 19.46   &  4.16 & 12.43       \\
                           & +adv    & \underline{10.62} & 19.43 & 8.16  & \underline{20.79}   & \underline{5.57}  & 12.91       \\
                           & +aware   & 10.37 & \underline{21.70} & \underline{8.25} & 20.07   & 5.26  & \underline{13.13}       \\
\midrule                        
\multirow{3}{*}{\ddec}      & MDBL         & 9.21  & \underline{12.39} & 6.64 & \underline{19.56}   &  3.27     & \underline{10.21}      \\
                           & +adv    & \underline{9.78}  & 11.01 & 6.76 & 18.03   & \underline{3.94}  & 9.90       \\
                           & +aware   & 9.51  & 12.25 & \underline{7.00}  & 18.46   & 3.41 & 10.13     \\                           
\midrule
\midrule  
\multirow{3}{*}{\shortstack[l]{\tenc\\ \denc}} & MDML       & 11.98 & 22.64 & 8.08  & 21.05   & 8.63      & 14.48 \\
                           & +adv   & \underline{12.11} & \textbf{\underline{23.43}} & \textbf{\underline{9.26}}  & \underline{21.59}   & 8.16      & 14.91 \\
                           & +aware & 11.82 & 23.07 & 9.08  & 21.54   & \underline{9.51}      & \underline{15.00} \\
\midrule
\multirow{3}{*}{\shortstack[l]{\tdec\\ \denc}} & MDML       & 10.57 & 18.32 & 6.87  & 20.04   & 10.25     & 13.21 \\
                           & +adv   & \underline{11.36} & \underline{22.69} & 7.13  & \underline{20.88}   & 9.44      & 14.30 \\
                           & +aware & 11.25  & 21.94 & \underline{8.73}  & 20.69   & \underline{10.34}     & \underline{14.59} \\
\midrule
\multirow{3}{*}{\shortstack[l]{\tdec\\ \ddec}} & MDML       & \underline{5.21}  & 17.56 & 4.53  & 9.12    & 4.36      & 8.16  \\
                           & +adv   & 2.25  & 17.47 & \underline{4.89}  & \underline{12.41}   & \underline{5.18}      & \underline{8.44}  \\
                           & +aware & 3.37  & \underline{18.85} & 4.23  & 9.35    & 4.14      & 7.99 \\
\midrule
\multirow{3}{*}{\shortstack[l]{\tenc\\ \ddec}} & MDML       & 9.39 & 21.29 & 8.28  & 22.06   & 9.54      & 14.11 \\
                           & +adv   & 10.84 & \underline{22.92} & \underline{8.45}  & 22.27   & 9.10      & 14.72 \\
                           & +aware & \textbf{\underline{12.29}}  & 22.70 & 8.39  & \textbf{\underline{22.62}}   & \textbf{\underline{10.75}}     & \textbf{\underline{15.35}} \\
\bottomrule                         
\end{tabular}}
\caption{BLEU score of En$\rightarrow$Cs translation
on leave-out domains for multi-domain multilingual (MDML) models and multi-domain bilingual (MDBL) models. {+adv and +aware denote MDML models trained with domain-agnostic or domain-aware auxiliary tasks, respectively.}
The best score on each domain overall and within each tagging group are marked in \textbf{bold} and \underline{underline} respectively.}
\label{tab:multidomain}
\end{center}
\vspace{-2mm}
\end{table*}

\begin{table*}[]
\begin{center}
\scalebox{1}{
\begin{tabular}{l|ll|rrrrrr}
\toprule
 & & \multicolumn{1}{c}{\textsc{Law}} & \multicolumn{1}{c}{\textsc{IT}} & \multicolumn{1}{c}{\textsc{Kor}} & \multicolumn{1}{c}{\textsc{Med}} & \multicolumn{1}{c}{\textsc{Sub}} & \multicolumn{1}{c}{AVG}  \\
\midrule \midrule
\multirow{6}{*}{\rotatebox[origin=c]{90}{\parbox[c]{1.5cm}{\centering T-ENC}}} & \multirow{2}{*}{(I)}  
& SDML &\textbf{49.21} & \textbf{41.63} & \textbf{32.33} & \textbf{51.84} & \textbf{32.01} & \textbf{41.40}
                         \\
& & MDML & 45.87 & 35.76 & 29.01 & 47.30 & 28.30 & 37.25 \\
\cline{2-9}
                       & \multirow{2}{*}{(II)} & SDML & 1.98 & 13.29 & 3.03 & 12.57 & 3.11 & 6.80 \\
                       & & MDML & \textbf{23.40} & \textbf{27.27} & \textbf{13.29} & \textbf{31.19} & \textbf{13.44} & \textbf{21.72} \\
\cline{2-9}                  
                       & \multirow{2}{*}{(III)} & SDML & 2.68 & 14.89 & 4.26 & 11.70 & 5.10 & 7.73 \\
                       & & MDML & \textbf{5.07} & \textbf{15.32} & \textbf{6.26} & \textbf{12.87} & \textbf{6.85} & \textbf{9.27} \\
\midrule \midrule
\multirow{6}{*}{\rotatebox[origin=c]{90}{\parbox[c]{1.5cm}{\centering T-DEC}}} & \multirow{2}{*}{(I)} & SDML  & \textbf{48.42} & \textbf{41.36} & \textbf{29.50} & \textbf{54.00} & \textbf{31.88} & \textbf{41.03} \\
& & MDML & 44.48 & 30.43 & 28.53 & 45.75 & 27.99 & 35.44 \\
\cline{2-9}
                       & \multirow{2}{*}{(II)} & SDML & 2.07  & 14.01 & 3.95 & 14.54 & 4.36 & 7.79 \\
& & MDML & \textbf{21.73} & \textbf{28.84} & \textbf{13.77} & \textbf{29.18} & \textbf{13.65} & \textbf{21.43} \\     
\cline{2-9}
 & \multirow{2}{*}{(III)} & SDML & 2.66 & 15.37 & 4.38 & 12.94 & 5.44 & 8.16 \\
& & MDML & \textbf{14.57} & \textbf{15.57} & \textbf{14.39} & \textbf{20.52} & \textbf{8.61} & \textbf{14.73} \\     
\bottomrule
\end{tabular}
}
\caption{Average BLEU scores of single-domain multilingual (SDML) and multi-domain multilingual (MDML) on the leave-out domains for three groups: (I) the three seen high-resource language pairs (En-De, En-Fr, De-Fr); (II) the two low-resource language pairs which are seen by MDML but unseen to SDML (En-Cs, En-Pl); and (III) the other five unseen language pairs.
}
\label{tab:sdml}
\end{center}
\end{table*}
\begin{table*}[]
\begin{center}
\scalebox{1}{
\begin{tabular}{llrrrrr|r}
\toprule
              & model      & \textsc{Law} & \textsc{IT} & \textsc{Kor} & \textsc{Med} & \textsc{Sub} & AVG \\
\midrule
\midrule
\multicolumn{2}{c}{Adapter-based} & 23.02 & 29.37 & 19.52 & 28.87 & 15.51 & 23.26 \\
\midrule
\multirow{3}{*}{\tenc}       & MDML       & 23.09                   & 27.86                  & 22.83                     & 34.19                       & 17.88                         & 25.17                   \\
 & +adv   & 28.74                   & 31.14                  & 25.68                     & 40.08                       & 18.89                         & 28.91                   \\
                             & +aware & \underline{31.56}                   & \textbf{32.00}                  & \underline{26.63}                     & \textbf{40.88 }                      & \underline{19.62}                         & \textbf{30.14}                   \\

\midrule
\multirow{3}{*}{\shortstack[l]{\tenc\\ \denc}} & MDML       & 21.14                   & 26.92                  & 21.84                     & 34.61                       & 17.31                         & 24.36                   \\
                             & +adv   & 26.09                   & \underline{31.91}                  & 26.09                     & 40.72                       & 19.67                         & 28.90                   \\
                             & +aware & \underline{27.10}                   & 31.85                  & \underline{26.40}                     & \underline{40.77}                       & \textbf{20.03}                         & \underline{29.23}                   \\
\midrule
\multirow{3}{*}{\shortstack[l]{\tenc\\ \ddec}} & MDML       & 20.04                   & 24.71                  & 19.57                     & 30.77                       & 15.43                         & 22.10                   \\
                             & +adv   & 30.14                   & 31.26                  & 26.03                     & 41.04                       & 18.68                         & 29.43                   \\
                             & +aware & \textbf{31.61}                   & \underline{31.49}                  & \underline{26.56}                     & \underline{40.84}                       & \underline{19.20}                         & \underline{29.94}                   \\
\midrule
\multirow{3}{*}{\tdec}       & MDML       & 25.92                   & 26.81                  & 20.49                     & 34.34                       & 16.56                         & 24.82                   \\

     & +adv   & \underline{29.64}                   & 30.91                  & 26.22                     & 40.31                       & 18.62                         & 29.14                   \\
                             & +aware & 29.45                   & \underline{31.54}                  & \textbf{26.81}                     & \underline{40.82}                       & \underline{19.01}                         & \underline{29.52}                   \\
\midrule
\multirow{3}{*}{\shortstack[l]{\tdec\\ \denc}} & MDML       & 24.89                   & 27.23                  & 20.97                     & 34.83                       & 16.85                         & 24.95                   \\
                             & +adv   & 27.66                   & 31.02                  & 25.54                     & 39.55                       & 19.02                         & 28.56                   \\
                             & +aware & \underline{28.15}                   & \underline{31.31}                  & \underline{25.77}                     & \underline{40.27}                       & \underline{19.56}                         & \underline{29.01}                   \\
\midrule
\multirow{3}{*}{\shortstack[l]{\tdec\\ \ddec}} & MDML       & \underline{21.24}                   & 19.24                  & \underline{16.06}                     & \underline{27.42}                       & \underline{12.01}                         & \underline{19.19}                   \\
                             & +adv   & 20.58                   & \underline{20.60}                  & 11.75                     & 24.09                       & 11.40                         & 17.68                   \\
                             & +aware & 13.91                   & 17.75                  & 9.49                      & 20.97                       & 9.74                          & 14.37                  \\
\bottomrule
\end{tabular}}
\caption{Average BLEU score of MDML-NMT models on each domain across all five leave-one-out scenarios and 20 {(seen and unseen)} translation tasks. 
The best score on each domain overall and within each tagging group are marked in \textbf{bold} and \underline{underline} respectively.
}
\label{tab:multilingual}
\end{center}
\end{table*}

\begin{table*}[]
\begin{center}
\scalebox{0.82}{   
\begin{tabular}{llrrrrrr|rrrrrr}
\toprule
                                   &                           & \multicolumn{6}{c}{seen (10)}                                                                               & \multicolumn{6}{|c}{unseen (zero-shot) (10)}    \\
                        
\multicolumn{1}{c}{} & \multicolumn{1}{c}{ } & \textsc{Law} & \textsc{IT} & \textsc{Kor} & \textsc{Med} & \textsc{Sub} & AVG & \textsc{Law} & \textsc{IT} & \textsc{Kor} & \textsc{Med} & \textsc{Sub} & AVG \\

\midrule
\midrule
\multicolumn{2}{c}{Adapter-based} & 43.38 & 36.15 & 26.92 & 50.06 & 27.27 & 36.76 & 10.97 & 23.23 & 27.69 & 35.05 & 12.16 & 21.82\\
\midrule
D-ENC & multi-domain & \\
D-DEC & multi-domain & \\
\midrule
\multirow{3}{*}{\tenc}             & MDML & 43.07 & 36.66 & 26.32 & 49.37 & 26.28 & 36.34 & 4.16 & 21.07 & 23.45 & 22.67 & 11.12 & 16.49 \\
                                  & +adv & 42.83 & 35.73 & 26.60 & 49.04 & 25.46 & 35.93 & 17.32 & 28.73 & 28.58 & 35.38 & 14.17 & 24.84 \\
                                  & +aware   & 43.36 & 36.54 & 27.08 & 49.84 & 25.97 & 36.56 & 22.88 & 29.64 & 29.91 & 36.62 & 15.38 & 26.88 \\
\multirow{3}{*}{\shortstack[l]{\tenc\\ \denc}}       & MDML  & 36.02 & 36.50 & 22.29 & 43.00 & 22.48 & 32.06 & 7.01 & 26.46 & 22.79 & 28.34 & 12.28 & 19.38 \\
                                  & +adv      & 43.08 & 36.41 & 26.76 & 49.46 & 25.99 & 36.34 & 10.63 & 29.77 & 28.31 & 36.49 & 15.02 & 24.05 \\
                                  & +aware & 43.41 & 36.71 & 27.02 & 49.74 & 26.15 & 36.61 & 12.95 & 29.20 & 29.25 & 36.64 & 15.37 & 24.68 \\
\multirow{3}{*}{\shortstack[l]{\tenc\\ \ddec}}       & MDML  & 36.06 & 36.32 & 22.24 & 42.74 & 21.86 & 31.85 & 3.87 & 21.95 & 19.36 & 21.10 & 9.92 & 15.24 \\
                                  & +adv  & 42.80 & 35.67 & 26.66 & 49.10 & 25.23 & 35.89 & 20.29 & 29.92 & 29.45 & 37.91 & 14.79 & 26.47\\
                                  & +aware & 43.53 & 36.50 & 27.39 & 49.94 & 25.84 & 36.64 & 23.20 & 29.72 & 29.82 & 36.98 & 15.32 & 27.01 \\
\midrule
\multirow{3}{*}{\tdec}             & MDML & 35.30 & 35.45 & 21.60 & 42.39 & 21.68 & 31.28 & 17.02 & 27.49 & 20.37 & 28.55 & 12.00 & 21.08 \\
                                  & +adv & 42.76 & 35.95 & 26.90 & 49.36 & 25.56 & 36.10 & 18.73 & 28.04 & 29.25 & 36.22 & 13.33 & 25.11 \\
                                  & +aware & 43.22 & 36.28 & 27.33 & 49.70 & 25.72 & 36.45 & 17.84 & 29.02 & 30.25 & 36.75 & 14.23 & 25.62 \\
\midrule
\multirow{3}{*}{\shortstack[l]{\tdec\\ \denc}}       & MDML  & 35.94 & 35.74 & 21.64 & 42.31 & 22.11 & 31.55 & 17.00 & 28.23 & 21.68 & 29.42 & 11.60 & 21.59 \\
                                  & +adv  & 42.08 & 35.45 & 26.85 & 48.55 & 25.41 & 35.67 & 15.87 & 28.49 & 29.12 & 34.95 & 13.71 & 24.43 \\
                                  & +aware  & 43.12 & 35.62 & 25.98 & 48.69 & 25.90 & 35.86 & 16.02 & 28.99 & 28.40 & 36.31 & 14.62 & 24.87 \\
\midrule
\multirow{3}{*}{\shortstack[l]{\tdec\\ \ddec}}       & MDML & 33.99 & 32.09 & 20.85 & 39.35 & 18.75 & 29.01 & 9.10 & 13.17 & 11.46 & 16.37 & 4.82 & 10.98 \\
                                    & +adv  & 35.67 & 27.11 & 19.13 & 35.53 & 17.76 & 27.04 & 6.92 & 15.26 & 4.74 & 14.39 & 5.75 & 9.41 \\
                                  & +aware   & 26.89 & 23.03 & 17.07 & 31.83 & 16.26 & 23.02 & 3.14 & 13.03 & 2.58 & 10.93 & 3.92 & 6.72 \\
\bottomrule                                   
\end{tabular}
}
\caption{Average BLEU score on other domains, i.e. excluding the leave-out domains, for different translation tasks. We categorise 20 translation direction into \emph{seen} where the translation direction in which training data are available, otherwise \emph{unseen}. The number in parentheses shows how many translation directions in the corresponding category.
}
\label{tab:other-domain}
\end{center}
\end{table*}

\end{document}